\documentclass[letterpaper, 10 pt, conference]{ieeeconf}  

\IEEEoverridecommandlockouts                              

\usepackage{graphicx} 

\usepackage{subcaption}
\usepackage{amsmath}
\usepackage{amsfonts}
\usepackage{amssymb}
\usepackage{multirow}

\usepackage{xcolor}

\newtheorem{dfn}{Definition}
\newtheorem{prop}{Proposition}

\definecolor{bleudefrance}{rgb}{0.19, 0.55, 0.91}

\title{\LARGE \bf
Designing Environments Conducive to Interpretable Robot Behavior}

\author{Anagha Kulkarni$^{1}$*, Sarath Sreedharan$^{1}$*, Sarah Keren$^{2}$, \\ Tathagata Chakraborti$^{3}$, David E. Smith$^{4}$, 
and Subbarao Kambhampati$^{1}$
\thanks{*Equal contribution}
\thanks{$^{1}$Anagha Kulkarni, Sarath Sreedhran and Subbarao Kambhampati are with Dept of Computer Science, Arizona State University
{\tt\small \{anaghak, ssreedh3, rao\}~ @asu.edu}}%
\thanks{$^{2}$Sarah Keren is with the School of Engineering and Applied Sciences, 
Harvard University, and Center for Research on Computation and Society, Harvard University
{\tt\small skeren@seas.harvard.edu}}%
\thanks{$^{3}$Tathagata Chakraborti is with IBM Research
{\tt\small tchakra2@ibm.com}}%
\thanks{$^{4}$David Smith can be reached at 
{\tt\small david.smith@psresearch.xyz}}%
}

\begin{document}

\maketitle
\thispagestyle{empty}
\pagestyle{empty}

\begin{abstract}

Designing robots capable of generating interpretable behavior is essential for effective human-robot collaboration. 
This requires robots to be able to generate behavior that aligns with human expectations but exhibiting such behavior in arbitrary environments could be quite expensive for robots, and in some cases, the robot may not even be able to exhibit
expected behavior.
However, in structured environments (like warehouses, restaurants, etc.), it may be possible to design the environment so as to boost the interpretability of a robot's behavior or to shape the human's expectations of the robot's behavior. In this paper, we investigate the opportunities and limitations of environment design as a tool to promote a particular type of interpretable behavior -- known in the literature as explicable behavior. We formulate a novel environment design framework that considers design over multiple tasks and over a time horizon. In addition, we explore the longitudinal effect of explicable behavior and the trade-off that arises between the cost of design and the cost of generating explicable behavior over an extended time horizon. 


\end{abstract}

\section{Introduction}

As more and more autonomous robots are deployed into environments cohabited by humans, the robot's capability of acting in a manner that is interpretable to the humans is becoming crucial.
This is because uninterpretable robot behavior not only leads to increased cognitive load on the human but also leads to loss of trust in the robot's capabilities and, in the worst case, may lead to 
increased risk around the robot \cite{fan2008influence}.
One way for the robot to be interpretable involves making its behavior consistent with the human's expectations of it. However, the human's expectation may deviate from reality as they may have an incorrect mental model about the robot's beliefs and capabilities. With an inconsistent human mental model, the robot's optimal behavior might not be interpretable to the human. 
In such cases, the robot should be able to reason over the inconsistencies between the two models to either generate \emph{explicable behavior}, which is consistent with the human's expectations of its behavior \cite{exp-yu,explicable-anagha}, or, \emph{explain} its behavior with respect to the inconsistencies in the human's mental model \cite{explain,sreedharan2018hierarchical}. 

However, the environment in which the robot is operating may not always be conducive to explicable behavior and/or to communicating explanations. As a result, making its behavior explicable may be prohibitively expensive for the robot.  In addition, certain behaviors that are explicable with respect to the human's mental model may not be feasible for the robot. Fortunately, in highly structured settings, where the robot is expected to solve repetitive tasks (like in warehouses, factories, restaurants, etc.), it might be feasible to \emph{redesign the environment} in a way that improves explicability of the robot's behavior, given a set of tasks. This brings us to the notion of environment design which involves redesigning the environment to maximize (or minimize) some objective for the robot 
\cite{zhang2009general}. Thus, environment design can be used to boost the explicability of the robot's behavior, especially in settings that require solving repetitive tasks and a one-time environment design cost to boost explicable behavior might be preferable over the repetitive cost overhead of explicable behavior borne by the robot. While the problem of environment design for planning problems has been investigated under the umbrella of {\em goal and plan recognition design} \cite{keren2014goal,mirsky2019goal}, 
they only form a subset of interpretable behaviors studied in the
existing literature \cite{landscape}.
To the best of our knowledge, we are the first 
to explore the notion of environment design 
to maximize the explicability of a robot's behavior.


However, environment design alone may not be a panacea for explicability. For one, the design could be quite expensive, not only in terms of making the required environment changes but also in terms of limiting the capabilities of the robot. Moreover, in many cases, there may not be a single set of design modifications that will work for a given set of tasks. For instance, designing a robot with wheels for efficient navigation on the floor will not optimize the robot's motion up a stairwell. 
This means, to achieve truly effective synergy with autonomous robots in a shared space, we need a greater synthesis of environment design and human-aware behavior generation. 
This leads us to investigate a novel optimization space, that requires trading off one-time (but potentially expensive) design changes, against repetitive costs borne by the robot to exhibit explicable behavior.

\begin{figure*}[!t]
\centering
\begin{subfigure}{\columnwidth} 
\centering
\includegraphics[width=\columnwidth]{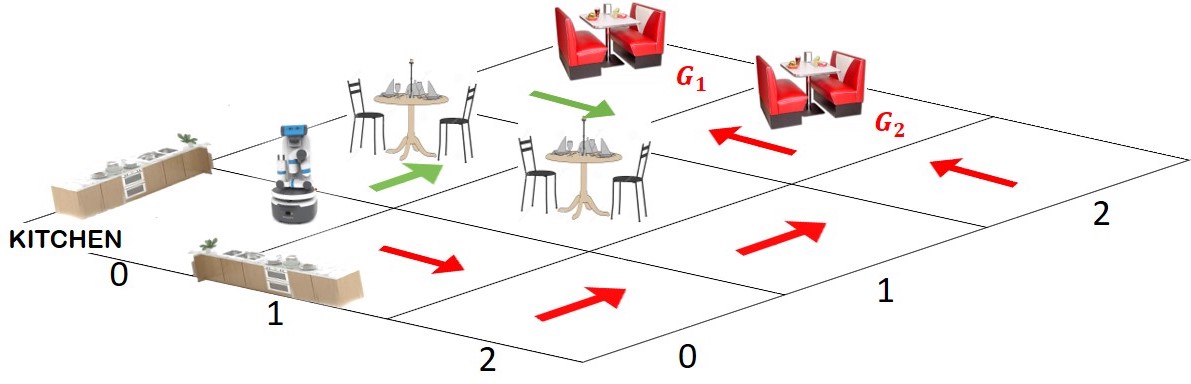}
\caption{Explicable behavior is costlier without design.}
\label{fig:ex1}
\end{subfigure} 
\hfill
\begin{subfigure}{\columnwidth}
\centering
\includegraphics[width=\columnwidth]{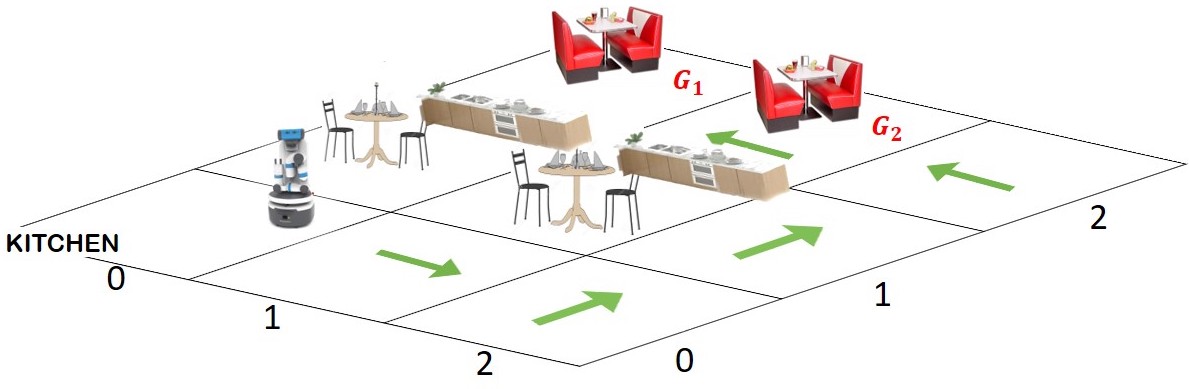}
\caption{Optimal behavior is explicable with design.}
\label{fig:ex2}
\end{subfigure}
\caption{
Use of environment design to improve the explicability of a robot's behavior in a shared environment.
}
\label{fig:example2}
\end{figure*}

The main contributions of our paper are as follows:
\begin{enumerate}
    \item We propose a new design framework that:
    \begin{enumerate}
        \item balances the cost of modifying the environment with the cost of inexplicability of a robot's behavior given the human's mental model,
        \item optimizes this objective given a set of tasks over a time horizon.
    \end{enumerate}
    \item Our work is the first to model the longitudinal aspect of explicable behavior, which captures the human's tolerance to inexplicability resulting from repetitive execution of tasks over a time horizon. 
    While this has been an issue with existing formulations of explicable behavior, the longitudinal impact of inexplicability becomes especially critical in the context of environment design which affects agents more permanently.   
    \item We leverage a planning compilation \cite{exact} to generate the most explicable plan for a task in a given environment and explore its theoretical properties. 
    \item Through empirical evaluation and demonstration of our approach in a simulated domain, we examine the properties of our optimization criterion and the various trade-offs that result from it. 
\end{enumerate}

\subsection{Motivating Example}

Consider a restaurant with a robot server (Figure \ref{fig:ex1}). 
Let $G_1$ and $G_2$ represent the robot's possible goals of serving the two booths: it travels between the kitchen and the two booths. 
The observers consist of customers at the restaurant. 
Given the position of the kitchen, the observers may have expectations on the route taken by the robot. However, unbeknownst to the observers, the robot can not traverse between the two tables and can only take the route around the tables. Therefore, the path marked in red is the cheapest path for the robot but the observers expect the robot to take the path marked in green in Figure \ref{fig:ex1}. 

In this environment, there is no way for the robot to behave as per the human's expectations. Applying environment design provides us with alternatives. For example, the designer could choose to build two barriers as shown in Figure \ref{fig:ex2}. With these barriers in place, the humans would expect the robot to follow the path highlighted in green. However, whether it is preferable to perform environment modifications or to bear the impact of inexplicable behavior depends on the cost of changing the environment versus the cost of inexplicability caused by the behavior.
In the rest of the paper, we will explore the details of this trade-off.

\section{Background}

\label{sec:background}

We consider two agents: a robot and a human observer. 
In this section, we introduce the notion of generating explicable behavior and the problem of environment design, with respect to these two agents.

\subsection{Planning}

A planning problem can be defined as a tuple $\mathcal{P}= \langle \mathcal{F}, \mathcal{A}, \mathcal{I}, \mathcal{G}, c \rangle $, where $\mathcal{F}$, is a set of fluents, $\mathcal{A}$, is a set of actions, and $c$ is the cost for each action. A state $s$ of the world is an instantiation of all fluents in $\mathcal{F}$. Let $\mathcal{S}$ be the set of states. $\mathcal{I} \in \mathcal{S}$ is the initial state. 
$\mathcal{G}$ is the goal where a subset of fluents in $\mathcal{F}$ are instantiated. 
Each action $a \in \mathcal{A}$ is a tuple of the form $\langle pre(a), add(a), del(a) \rangle$ where $pre(a) \subseteq \mathcal{F}$ is a set of preconditions, $add(a) \subseteq \mathcal{F}$ is a set of add effects and $del(a) \subseteq \mathcal{F}$ is a set of delete effects of action $a$. 
The transition function $\Gamma_\mathcal{P}(\cdot)$ is given by $\Gamma_\mathcal{P}(s, a) \models \bot$ if $s \not\models pre(a)$; else $\Gamma_\mathcal{P}(s, a) \models s \cup add(a) \setminus del(a)$.
The solution to $\mathcal{P}$ is a \emph{plan} or a sequence of actions $\pi = \langle a_1, a_2, \ldots, a_n \rangle$, such that, $\Gamma_\mathcal{P}(\mathcal{I},\pi) \models \mathcal{G}$, i.e., starting from the initial state and sequentially executing the actions results in the robot achieving the goal. The cost of the plan, $c(\pi)$, is a sum of the cost of all the actions in it, $c(\pi) = \sum_{a_i\in\pi}c(a_i)$.

\subsection{Explicability}

Let $\mathcal{P}_R = \langle \mathcal{F}, \mathcal{A}_R, \mathcal{I}_R, \mathcal{G}_R, c_R \rangle $ be the robot's model captured as a planning problem. The need for generating explicable behavior arises because the robot's planning model is different from the human's mental model of it. The difference can be in terms of a set of actions, the initial state or goal of the robot.  Thus an explicable planning problem is defined as $\mathcal{P}_{Exp}= \langle \mathcal{P}_R, \mathcal{P}_H, \delta_{\mathcal{P}_H}\rangle$, where $\mathcal{P}_H = \langle \mathcal{F}, \mathcal{A}_H, \mathcal{I}_H, \mathcal{G}_H, c_H \rangle $ represents the human's mental model of the robot model, and $\delta_{\mathcal{P}_H}$ is a distance function used by the human to compute the explicability of a plan. We assume the human mental model as an input. This is usually the case when any product is deployed and developers capture a generic user model which can be learned from prior interactions. In this work, we only focus on the reasoning aspects once we have the model, rather than focusing on the acquisition of such a model. 
Let $\Pi^*_{\mathcal{P}_H}$ represent the set of expected plans with respect to $\mathcal{P}_H$. Here, $\Pi^*_{\mathcal{P}_H}$ captures the notion of the human’s preference on the plans feasible in their mental model.
A valid plan that solves $\mathcal{P}_R$ can exist anywhere on the spectrum of inexplicability from high to low. 

\vspace{5pt}
\begin{dfn}
The \textbf{inexplicability score}, $\mathcal{IE}(\cdot, \cdot, \cdot)$, of the robot's plan $\pi_R$ that solves $\mathcal{P}_{R}$ is defined as follows for the human's mental model $\mathcal{P}_{H}$ and a distance function $\delta_{\mathcal{P}_H}(\cdot, \cdot)$:
\begin{equation}
\mathcal{IE}(\pi_R, \mathcal{P}_{H}, \delta_{\mathcal{P}_{H}}) = \min_{\pi_H \in \Pi^*_{\mathcal{P}_{H}}} {\delta_{\mathcal{P}_H}(\pi_R, \pi_H)}
\label{eq:exp_score}
\end{equation}
where $\delta_{\mathcal{P}_H}(\cdot, \cdot)$ is a distance function that assesses the difference between the two plans $\pi_R$ and $\pi_H$.
\end{dfn}

The robot plans to minimize the inexplicability score in the human's mental model.
We will use the notation $\Pi^*_{\mathcal{IE}(\cdot, \mathcal{P}_{H}, \delta_{\mathcal{P}_{H}})}$ (in the absence of the parameter $\pi_R$) to refer to the set of plans in the robot's model with the lowest inexplicability score, and $\mathcal{IE}_{min}(\mathcal{P}_{Exp})$ to represent the lowest inexplicability score associated with the set. Further, let $f_{Exp}$ be the decision function used by the explicable robot: $f_{Exp}(\mathcal{P}_{Exp})$ represents the cheapest plan that minimizes the inexplicability score, i.e. $f_{Exp}(\mathcal{P}_{Exp}) \in \Pi^*_{\mathcal{IE}(\cdot, \mathcal{P}_{H}, \delta_{\mathcal{P}_{H}})}$ and $\lnot\exists \pi': \pi' \in \Pi^*_{\mathcal{IE}(\cdot, \mathcal{P}_{H}, \delta_{\mathcal{P}_{H}})}$ such that $c_R(\pi') < c_R(f_{Exp}(\mathcal{P}_{Exp}))$.

\subsection{Environment Design} An environment design problem \cite{zhang2009general} takes as input the initial environment configuration along with a set of available modifications and computes a subset of modifications that can be applied to the initial environment to derive a new environment in which a desired objective is optimized. 

We consider $\mathcal{P}^0_R = \langle \mathcal{F}^0, \mathcal{A}^0_R, \mathcal{I}^0_R, \mathcal{G}^0_R, c^0_R \rangle$ as the initial environment
and $\rho_R$ as the set of valid configurations of that environment: $\mathcal{P}^0_R \in \rho_R$. Let $\mathcal{O}$ be an arbitrary metric that needs to be optimized with environment design, i.e a planning model with lower value for $\mathcal{O}$ is preferred. 
A design problem (adapted from \cite{zhang2009general}) is a tuple $\langle \mathcal{P}^0_R, \Delta,  \Lambda_R, C, \mathcal{O} \rangle$ where, $\Delta$ is the set of all modifications, $\Lambda_R: \rho_R \times 2^\Delta \rightarrow \rho_R$ is the model transition function that specifies the resulting model after applying a subset of modifications to the existing model, $C: \Delta \rightarrow \mathbb{R}$ is the cost function that maps each design choice to its cost. The modifications are independent of each other and their costs are additive. We will overload the notation and use $C$ as the cost function for a subset of modifications as well, i.e. $C(\xi) = \sum_{\xi_i \in \xi} C(\xi)$. 



The set of possible modifications could include modifications to the state space, action preconditions, action effects, action costs, initial state and goal. In general, the space of design modifications, which are an input to our system, may also involve modifications to the robot itself (since the robot is part of the environment that is being modified).
An optimal solution to a design problem identifies the subset of design modifications, $\xi$, that minimizes the following objective  
consisting of the cost of modifications and the metric $\mathcal{O}$: ~$\min \mathcal{O}(\Lambda_R(\mathcal{P}^0_R, \xi)),~ C(\xi)$.

\section{Design for Explicability}

In this framework, we not only discuss the problem of environment design with respect to explicability but also in the context of \textbf{(1)} a set of tasks that the robot has to perform in the environment, and \textbf{(2)} over the lifetime of the tasks i.e. the time horizon over which the robot is expected to repeat the execution of the given set of tasks. 
These considerations add an additional dimension to the environment design problem since the design will have lasting effects
on the robot's behavior. In the following, we will first introduce the design problem for a single explicable planning problem, then extend it to a set of explicable planning problems and lastly extend it over a time horizon.

\subsection{Design for a Single Explicable Problem}

In the design problem for explicability, the inexplicability score becomes the metric that we want to optimize for. That is we want to find an environment design such that the inexplicability score is reduced in the new environment.  This problem can be defined as follows:

\vspace{5pt}
\begin{dfn}
The design problem for explicability is a tuple, $\mathcal{DP}_{Exp} = \langle \mathcal{P}^0_{Exp}, \Delta, \Lambda_{Exp}, C, \mathcal{IE}_{min} \rangle$, where:
\begin{itemize}
\item $\mathcal{P}^0_{Exp} \in \rho_{Exp}$ is the initial configuration of the explicable planning problem, where $\rho_{Exp}$ represents the set of valid configurations for $\mathcal{P}_{Exp}$. 
\item $\Delta$ is the set of available design modifications. The space of all possible modifications is the power-set $2^\Delta$.
\item $\Lambda_{Exp}: \rho_{Exp} \times 2^\Delta \rightarrow \rho_{Exp}$ is the transition function over the explicable planning problem, which gives an updated problem after applying the modifications.  
\item $C$ is the additive cost associated with each design in $\Delta$.
\item $\mathcal{IE}_{min}:\rho_{Exp} \rightarrow \mathbb{R}$ is the minimum possible inexplicability score in a configuration, i.e. the inexplicability score associated with the most explicable plan.
\end{itemize}
\end{dfn}

\vspace{5pt}
With respect to our motivating example in Figure \ref{fig:ex1}, $\mathcal{DP}_{Exp}$ is the problem of designing the environment to improve the robot's explicability given its task of serving every new customer at a booth (say $G_1$) only once. The optimal solution to $\mathcal{DP}_{Exp}$ involves finding a configuration which minimizes the minimum inexplicability score.
We also need to take into account an additional optimization metric which is the effect of design modifications on the robot's plan cost. That is, we need to examine to what extent the decrease in inexplicability is coming at the robot's expense. For instance,
if you confine the robot to a cage so that it cannot move, its behavior becomes completely and trivially explicable, but the cost of achieving its goals goes to infinity. 

\vspace{5pt}
\begin{dfn}
An \textbf{optimal solution} to $\mathcal{DP}_{Exp}$, is a subset of modifications $\xi^*$ that minimizes the following: 
\begin{align}
\min~ \mathcal{IE}_{min}(~\mathcal{P}^*_{Exp}),~ C(\xi^*), c_R(f_{Exp}(\mathcal{P}^*_{Exp})) 
\label{eq:sol}
\end{align}
where $\mathcal{P}^*_{Exp} = \Lambda_{Exp}(\mathcal{P}^0_{Exp}, \xi^*)$ is the final modified explicable planning problem, 
$\mathcal{IE}_{min}(\cdot)$ represents the minimum possible inexplicability score for a given configuration, 
$C(\xi^*)$ denotes the cost of the design modifications and $c_R(f_{Exp}(\mathcal{P}^*_{Exp}))$ is the cost of the cheapest most explicable plan in a configuration.
\end{dfn}

\subsection{Design for Multiple Explicable Problems}

We will now show how $\mathcal{DP}_{Exp}$ evolves when there are multiple explicable planning problems in the environment that the robot needs to solve. When there are multiple tasks there may not exist a single set of design modifications that may benefit all the tasks. In such cases, a solution might involve performing design modifications that benefit some subset of the tasks while allowing the robot to act explicably with respect to the remaining set of tasks. Let there be $k$ explicable planning problems, given by the set $\mathbf{P}_{Exp} = \{ \langle \mathcal{P}_{R}(0), \mathcal{P}_{H}(0), \delta_{\mathcal{P}_{H}(0)}  \rangle, \ldots, \langle \mathcal{P}_{R}(k), \mathcal{P}_{H}(k), \delta_{\mathcal{P}_{H}(k)} \rangle\}$, with a categorical probability distribution $\mathcal{D}$ over the problems. We use $\mathcal{P}_{Exp}(i) \in \mathbf{P}_{Exp}$ to denote the $i^{th}$ explicable planning problem. These $k$ explicable problems may differ in terms of their initial state and goal conditions.
Now the design problem can be defined as:
\begin{align}
\mathcal{DP}_{Exp,\mathcal{D}} = \langle \mathbf{P}^0_{Exp}, \mathcal{D}, \Delta, \Lambda_{Exp}, C, \mathcal{IE}_{min,\mathcal{D}}  \rangle,
\end{align}
where $\mathbf{P}^0_{Exp}$, is the set of planning tasks in the initial environment configuration, $\mathcal{IE}_{min,\mathcal{D}}$ is a function that computes the minimum possible inexplicability score in a given environment configuration by taking the expectation over the minimum inexplicability score for each explicable planning problem, 
i.e., $\mathcal{IE}_{min,\mathcal{D}}(\mathbf{P}_{Exp}) = \mathbb{E}[\mathcal{IE}_{min}(\mathcal{P}_{Exp})]$, where $\mathcal{P}_{Exp} \sim \mathcal{D}$. With respect to our running example, $\mathcal{DP}_{Exp,\mathcal{D}}$ is the problem of designing the environment 
given the robot's task of serving every new customer only once at either of the booths ($G_1$, $G_2$) with probability given by  $\mathcal{D}$. 

The solution to $\mathcal{DP}_{Exp,\mathcal{D}}$ has to take into account the distribution over the set of explicable planning problems. Therefore the optimal solution is given by: 
\begin{align}
\min~ ~\mathcal{IE}_{min, \mathcal{D}}(~\mathbf{P}^*_{Exp}),~ C(\xi^*),~
\mathbb{E}[c_R(f_{Exp}(\mathcal{P}^*_{Exp}) )]
\end{align}

\noindent where $\mathcal{P}^*_{Exp} \sim \mathcal{D}$. 
A valid configuration minimizes the minimum possible inexplicability score, which involves 1) expectation over minimum inexplicability scores for each explicable planning problem; 2) the cost of the design modifications (these modifications are applied to each explicable planning problem); and 3) the expectation over the cheapest most explicable plan for each explicable planning problem.

\begin{figure}[!t]
\centering
\includegraphics[width=\columnwidth]{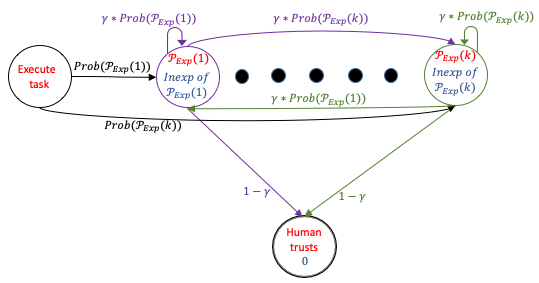}
\caption{Illustration of longitudinal impact on explicability. \emph{Prob} determines the probability associated with executing each task in $\mathbf{P}_{Exp}$. For each task, the reward is determined by the inexplicability score of that task. The probability of achieving this reward is determined by $\gamma$ $\times$  probability of executing that task. Additionally, with a probability $(1 - \gamma)$ the human
ignores the inexplicability of a task and the associated reward is given by an inexplicability score of 0.}
\label{fig:ex3}
\end{figure} 

\subsection{Longitudinal Impact on Explicable Behavior}

The process of applying design modifications to an environment makes more sense if the tasks are going to be performed repeatedly in the presence of a human (i.e. the robot does not have to bear the cost of being explicable repeatedly).
This has quite a different temporal characteristic in comparison to that of execution of one-time explicable behavior. For instance, design changes are associated with a one-time cost (i.e. the cost of applying those changes in the environment). On the other hand, if we are relying on the robot to execute explicable plans at the cost of foregoing optimal plans, then it needs to bear this cost multiple times in the presence of a human over the time horizon. 

We will use a discrete time formulation where the design problem is associated with a time horizon $\mathcal{T}$. At each time step, one of the $k$ explicable planning problems is chosen. Now the design problem can be defined as:
\begin{align}
\mathcal{DP}_{Exp,\mathcal{D}, \mathcal{T}} = \langle \mathbf{P}^0_{Exp}, \mathcal{D}, \Delta, \Lambda_{Exp}, C, \mathcal{IE}_{min,\mathcal{D}}, \mathcal{T} \rangle
\end{align}

In our running example, $\mathcal{DP}_{Exp,\mathcal{D}, \mathcal{T}}$ is the problem of designing the environment given the robot's task of serving the same customer at either of the booths with a distribution $\mathcal{D}$ over a horizon $\mathcal{T}$.

In the past literature, the explicable behavior has been studied with respect to a single interaction with a human over a given task \cite{exp-yu,explicable-anagha}. However, we consider a time horizon, $\mathcal{T} > 1$, over which the robot's interaction with the human may be repeated multiple times for the same task. This means the human's expectations about the task can evolve over time. This may not be a problem if the robot's behavior aligns perfectly with the human's expectations. Although, if the robot's plan for a given task is associated with a non-zero inexplicability score, then the human is likely to be more surprised the very first time she notices the inexplicable behavior than she would be if she noticed the inexplicable behavior subsequent times.
As the task is performed over and over, the amount of surprise associated with the inexplicable behavior starts decreasing. In fact, there is a probability that the human may ignore the inexplicability of the robot's behavior after sufficient repetitions of the task. We incorporate this intuition by using discounting. 

Figure \ref{fig:ex3} illustrates the Markov reward process to represent the dynamics of this system. Let $(1 - \gamma)$ denote the probability that the human will ignore the inexplicability of the robot's plan, i.e, the reward will have inexplicability score 0. $\gamma$ times the probability of executing a task represents the probability that the reward will have the minimum inexplicability score associated with that task. Assuming $\gamma < 1$, the minimum possible inexplicability score for a set of explicable planning problems is:
\begin{align}
f_\mathcal{T}(\mathcal{IE}_{min,\mathcal{D}}(\mathbf{P}_{Exp})) 
&= \mathcal{IE}_{min,\mathcal{D}}(\mathbf{P}_{Exp}) \nonumber \\
& \ + \gamma * \mathcal{IE}_{min,\mathcal{D}}(\mathbf{P}_{Exp}) + \ldots + \nonumber \\
& \ \ \gamma^{T-1} * \mathcal{IE}_{min,\mathcal{D}}(\mathbf{P}_{Exp}) \nonumber \\
f_\mathcal{T}(\mathcal{IE}_{min,\mathcal{D}}(\mathbf{P}_{Exp})) 
&= \frac{1-\gamma^\mathcal{T}}{1-\gamma} * \mathcal{IE}_{min,\mathcal{D}}(\mathbf{P}_{Exp}) 
\end{align}



Thus the optimal solution to $\mathcal{DP}_{Exp,\mathcal{D}, \mathcal{T}}$ is given by: 
\begin{align}
\min~ ~f_\mathcal{T}(\mathcal{IE}_{min, \mathcal{D}}(\mathbf{P}^*_{Exp})),~ C(\xi^*),
\nonumber \\ 
~ \mathbb{E}[c_R(f_{Exp}(\mathcal{P}^*_{Exp}))] * \mathcal{T}
\end{align}
\noindent where, $\mathcal{P}^*_{Exp} \sim \mathcal{D}$. The optimal solution is a valid configuration that minimizes 1) the minimum possible inexplicability over the set of explicable planning problems given the human's tolerance to inexplicable behavior; 2) one-time cost of the design modifications; and 
3) the expectation over the cheapest most explicable plan for each explicable planning problem given a time horizon.
Note that, since the design cost is not discounted and we always make the design changes before the task is solved, there is never a reason to delay the design execution to future steps in the horizon. Instead it can be executed before the first time step.

\section{Solution Methodology}

We now discuss a solution strategy for our design problem when a cost-based distance function ($\delta^c_{\mathcal{P}_H}$) is used to determine the inexplicability of a plan. Given a plan $\pi$, such that, $\Gamma_{\mathcal{P}_R}(\mathcal{I}_R, \pi) \models \mathcal{G}_R$, 
the distance from an expected plan $\pi'$ in the human model is given as $\delta^c_{\mathcal{P}_H}(\pi, \pi') =$

\vspace{-10pt}
\begin{align}
\begin{cases} \exp(|c_H(\pi) - c_H(\pi')|), \quad \hfill if~ \Gamma_{\mathcal{P}_H}(\mathcal{I}_H, \pi) \models \mathcal{G}_H \\ 
\infty, \hfill otherwise \end{cases}
\end{align}

\vspace{5pt}
Here, we will use the set of plans that are optimal in the human's mental model as the expected plan set. This means that for calculating Equation \ref{eq:exp_score}, we do not require an additional minimization over the space of expected plans as every plan in the robot's model should be equidistant from every optimal plan in the human's mental model (and the distance is infinity if the current robot plan is not executable in the human's mental model). For brevity, we refer to any plan with infinite inexplicable score as being invalid for a problem in $\mathbf{P}_{Exp}$. Also, we assume that the actions in both the models have unit costs. That is, $c_H(\pi) = c_R(\pi)$ = $|\pi|$.

\vspace{5pt}
\begin{prop}
$\forall i \in {1, \ldots, k}, \pi, \pi' \in \Pi^*_{\mathcal{IE}(\cdot, \mathcal{P}_{H}(i), \delta_{\mathcal{P}_{H}(i)})}, \\ c_R(\pi) = c_R(\pi')$.
\end{prop}

\vspace{5pt}
The above proposition states that all plans in $\Pi^*_{\mathcal{IE}(\cdot, \mathcal{P}_{H}(i), \delta_{\mathcal{P}_{H}(i)})}$ have equal costs in $\mathcal{P}_R(i)$ due to the assumption of unit costs. Therefore, while calculating the value for the objective function of $\mathcal{DP}_{Exp,\mathcal{D}, \mathcal{T}}$, we can choose an arbitrary plan from $\Pi^*_{\mathcal{IE}(\cdot, \mathcal{P}_{H}(i), \delta_{\mathcal{P}_{H}(i)})}$ to calculate the term corresponding to the robot's cost. 



\subsection{Search for Optimal Design}
To find the optimal solution for $\mathcal{DP}_{Exp,\mathcal{D}, \mathcal{T}}$, we will perform a breadth-first search over the space of environment configurations that are achievable from the initial configuration through the application of the given set of modifications \cite{keren2018strong}. The performance of the search depends on the number of designs available. By choosing appropriate design strategies, significant scale up can be attained. 
Each search node is a valid environment configuration and the possible actions are the applicable designs. For simplicity, we convert the multi-objective optimization in Equation \ref{eq:sol} into a single objective as a linear combination of each term associated with a coefficients $\alpha$, $\beta$, and $\kappa$, respectively. The value of each node is decided by the aforementioned objective function. For each node, it is straightforward to calculate the design modification cost. However, in order to calculate the minimum inexplicability score and the robot's plan cost, we have to generate a plan that minimizes the inexplicability score for each explicable planning problem in that environment configuration. To achieve this, we compile the problem of generating the explicable plan to a classical planning problem. We will discuss this compilation in the following subsection. Essentially, our search has two loops: the outer loop which explores all valid environment configurations, and the inner loop which performs search in a valid environment configuration to find a plan that minimizes the inexplicability score. At the end of the search, the node with best value is chosen, and the corresponding set of design modifications, $\xi^*$, is output. 

One way to optimize our search over the space of environment configurations is to only consider the designs that are relevant to the actions in the optimal robot plans ($\Pi^*_{\mathbf{P}_R}$) and those in the human's expected plans ($\Pi^*_{\mathbf{P}_H}$) given the set of tasks. This can be implemented as a pruning strategy that prunes out designs that are not relevant to the actions.

\subsection{Compilation for Most Explicable Plan} 
We show that generating the most explicable plan for a $\mathcal{P}_{Exp} = \langle \mathcal{P}_R, \mathcal{P}_H, \delta_{\mathcal{P}_H} \rangle$ is the same as generating an optimal plan, $\pi^*_{mod}$, for a transformed planning problem $\mathcal{P}_{mod}$. To this end, we leverage the compilation used by \cite{exact} and present a simplified version.

\vspace{5pt}
\begin{dfn}
Given an explicable planning problem, $\mathcal{P}_{Exp} = \langle \mathcal{P}_R, \mathcal{P}_H, \delta_{\mathcal{P}_H} \rangle$, the transformed planning problem is $\mathcal{P}_{mod} = \langle \mathcal{F}_{mod}, \mathcal{A}_{mod}, \mathcal{I}_{mod}, \mathcal{G}_{mod}, c_{mod} \rangle$, where, 
\vspace{5pt}
\begin{itemize}
\item $\mathcal{F}_{mod}  = \mathcal{F}_R \cup \mathcal{F}_{H}$\\[-2ex]
\item For each $a_{mod} \in \mathcal{A}_{mod}$, $a_{mod} = \langle pre(a_{mod}),$ $add(a_{mod}),$ $del(a_{mod})\rangle$, where: \\[1ex]
\begin{math}
pre(a_{mod}) = \{ f_R |  f \in pre(a_R)\} \cup \{f_H | f \in pre(a_H) \} \\
add(a_{mod}) = \{ f_R |  f \in add(a_R)\} \cup \{f_H | f \in add(a_H) \} \\
del(a_{mod}) = \{ f_R |  f \in del(a_R) \} \cup \{f_H | f \in del(a_H) \}\end{math}
\vspace{5pt}
\item $\mathcal{I}_{mod}  = \{f_R |  f \in  \mathcal{I}_R \} \cup \{f_H | f \in \mathcal{I}_H \}$, and \\ 
$\mathcal{G}_{mod}  = \{f_R | f \in  \mathcal{G}_R \} \cup \{f_H | f \in \mathcal{G}_H \}$\\[-2ex]
\item For each $a_{mod} \in \mathcal{A}_{mod}$, $c_{mod}(a_{mod}) = c_{H}(a_H) =1$
\end{itemize}
\end{dfn}

\vspace{5pt}
We label the fluents with different subscripts to denote that we maintain two separate copies of fluents in the transformed planning problem: i.e., for every $f \in \mathcal{F}$, there is robot's fluent, $f_R \in \mathcal{F}_R$ and the human's belief about it, $f_H \in \mathcal{F}_H$. We assume there is a one to one mapping between the actions in the robot's model and those in the human's mental model, so there are two versions of each action. The action transformation ensures that an action is executable by the robot if and only if its preconditions are satisfied in both $\mathcal{P}_R$ and $\mathcal{P}_H$, and it produces effects consistent with both models.

\vspace{5pt}
\begin{prop}
The problem $\mathcal{P}_{mod}$ produces a plan that solves $\mathcal{P}_{Exp}$, so that the following properties hold:
\begin{itemize}
\item \textbf{Soundness} A plan $\pi_{mod}$ that solves $\mathcal{P}_{mod}$ is a valid solution for $\mathcal{P}_{Exp}$.\\[-2ex]
\item \textbf{Completeness} For every valid plan that solves $\mathcal{P}_{Exp}$, there is a corresponding valid plan that solves $\mathcal{P}_{mod}$.\\[-2ex]
\item \textbf{Optimality} A plan $\pi^*_{mod}$ that solves $\mathcal{P}_{mod}$ optimally is the most explicable plan for $\mathcal{P}_{Exp}$.
\end{itemize}
\end{prop}

\vspace{5pt}
\begin{proof}
The transformed planning problem has the union of the constraints imposed by both $\mathcal{P}_R$ and $\mathcal{P}_H$. Given a plan $\pi$, such that, $\Gamma_{\mathcal{P}_{mod}}(\mathcal{I}_{mod}, \pi) \models \mathcal{G}_{mod}$, by the definition of the compilation, we also have $\Gamma_{\mathcal{P}_{R}}(\mathcal{I}_{R}, \pi) \models \mathcal{G}_{R}$ and $\Gamma_{\mathcal{P}_{H}}(\mathcal{I}_{H}, \pi) \models \mathcal{G}_{H}$. Hence, a plan $\pi_{mod}$ that solves $\mathcal{P}_{mod}$ is a valid plan for $\mathcal{P}_{Exp}$.

From the definition of the inexplicability score for a plan $\pi_R$ which is a valid solution to $\mathcal{P}_{Exp}$, we know that $\Gamma_{\mathcal{P}_{H}}(\mathcal{I}_{H}, \pi_R) \models \mathcal{G}_{H}$. Such a plan $\pi_R$ solves both $\mathcal{P}_{R}$ and $\mathcal{P}_{H}$. Hence, $\pi_R$ will satisfy, $\Gamma_{\mathcal{P}_{mod}}(\mathcal{I}_{mod}, \pi_R) \models \mathcal{G}_{mod}$. Therefore, for every valid plan that solves $\mathcal{P}_{Exp}$, there exists a corresponding plan that solves $\mathcal{P}_{mod}$. 

Given $\mathcal{P}_{Exp}$, let $\pi'$ be the most explicable robot plan (the plan with lowest inexplicability score) which is not an optimal plan for $\mathcal{P}_{mod}$.
By definition of explicability, this means $\pi'$ must be a valid plan for both $\mathcal{P}_R$ and $\mathcal{P}_H$. Further, by the completeness property, we know that $\pi'$ must be a valid plan for $\mathcal{P}_{mod}$. This means that for a plan $\pi^*_{mod}$ optimal in $\mathcal{P}_{mod}$, we have $c_H(\pi^*_{mod}) < c_H(\pi')$ (since $\mathcal{P}_{mod}$ uses $c_H$). 
Hence, $|c_H(\pi^*_{mod}) - c_H^{*}| < |c_H(\pi') - c_H^{*}|$, where $c_H^*$ is the cost of an optimal plan in $\mathcal{P}_H$ (and we know $c_H^* \leq c_H(\pi^*_{mod})$ and $c_H^* \leq c_H(\pi')$). This means $\mathcal{IE}(\pi^*_{mod}, \mathcal{P}_H, \delta_{\mathcal{P}_H}) < \mathcal{IE}(\pi', \mathcal{P}_H, \delta_{\mathcal{P}_H})$. 
This contradicts the initial assertion, proving that there is a one to one correspondence between optimal plans for $\mathcal{P}_{mod}$ and $\Pi^*_{\mathcal{IE}(\cdot, \mathcal{P}_{H}, \delta_{\mathcal{P}_{H}})}$.
\end{proof}

\section{Evaluation}

We will now demonstrate how the explicability value and design cost of the optimal solution evolve when optimizing for a single problem, multiple problems and multiple problems with a time horizon using the running example. We will also evaluate the performance of our approach on 3 IPC (International Planning Competition) domains and discuss the interplay between explicability and plan cost.  


\subsection{Demonstration} We use our running example from Figure \ref{fig:ex1} to demonstrate how the design problem evolves. We constructed a domain where the robot had 3 actions: \textit{pick-up} and \textit{put-down} to serve the items on a tray and \textit{move} to navigate between the kitchen and the booths. Some grid cells are blocked due to the tables and the robot cannot pass through these: cell(0, 1) and cell(1, 1). Therefore, the following passages are blocked: cell(0, 0)-cell(0, 1), cell(0, 1)-cell(0, 2), cell(0, 1)-cell(1, 1), cell(1, 0)-cell(1, 1), cell(1, 1)-cell(1, 2), cell(1, 1)-cell(2, 1). We considered 6 designs, each consisting of putting a barrier at one of the 6 passages to indicate the inaccessibility to the human (i.e. the design space has $2^6$ possibilities). 

For the following parameters: $\alpha = 1$, $\beta = 30$, $\kappa = 0.25$ and $\gamma = 0.9$, we ran our algorithm for three settings: (a) single explicable problem for $\mathcal{T} = 1$, (b) multiple explicable problems for $\mathcal{T} = 1$, and (c) multiple explicable problems for $\mathcal{T} = 10$. As mentioned before, (a) involved serving a new customer at a booth (say $G_1$) only once, (b) involved serving a new customer only once at either of the booths with equal probability and (c) involved serving each customer at most 10 times at either of the booths with equal probability. We found that for settings (a) and (b) no design was chosen. This is because these settings are over a single time step and the cost of installing design modifications in the environment is higher than the amount of inexplicability caused by the robot ($\beta > \alpha$). On the other hand, for setting (c), the algorithm generated the design in Figure \ref{fig:ex2}, which makes the robot's roundabout path completely explicable to the customers.  

\subsection{Domain setup} We used three IPC domains for evaluation: \texttt{Blocksworld}, \texttt{IPC-Grid} and \texttt{Driverlog}. 
For each domain, we created two versions: the robot's domain and the human's domain. 
We generated 20 design problems for each domain, and each had 3 planning problems with uniform probability distribution. All the experiments were run on an Ubuntu workstation with 64G RAM.
We used Fast Downward with A* search and the \textit{lmcut} heuristic \cite{helmert2006fast} to solve the compiled planning problems. The variable parameters in our implementation are $\alpha$, $\beta$, $\kappa$ (coefficients associated with the terms in the objective function), $\gamma$ (discount factor) and $\mathcal{T}$ (time horizon). For all the domains we used actions and design modifications of unit cost. 

For \texttt{Blocksworld}, the robot's domain was the original IPC domain, and the human's domain assumed that the robot can pick up multiple blocks simultaneously. 
The set of allowed designs ensured that stacking for every block was preceded by picking the block up from the table. This would reduce the inexplicability for the human as the only block that would be stacked is the one that was picked up from the table before stacking. In practice, this may involve notifying the human about the new rule.
For \texttt{IPC-Grid}, the robot's domain was the original IPC domain and the human's domain assumed that diagonal movements were possible in the grid. 
We allowed design modifications that pruned diagonal actions. In actuality, this may involve notifying the human that diagonal actions are not possible at certain locations. 
For \texttt{Driverlog}, the robot's domain was the original IPC domain and the human's domain assumed that packages can be loaded and unloaded from the truck regardless of the location of the driver. 
We allowed modifications that required load and unload actions to occur only after a disembark action. This may again involve notifying the human about the new rules concerning load/unload actions. 


\setlength{\tabcolsep}{4pt}
\renewcommand{\arraystretch}{1.5}
\begin{table*}[!t] 
\centering
\resizebox{\textwidth}{!}{%
\begin{tabular}{ |l|c|c|c|c|c|c|c|c|c|c|c|c|c| }
\hline
\multirow{2}{*}{Domain} & \multirow{2}{*}{Horizon} & \multirow{2}{*}{Metrics} & \multirow{2}{*}{Design} &  \multicolumn{3}{c|}{Inexplicability} &
\multicolumn{3}{c|}{Plan Cost} &
\multicolumn{3}{c|}{Total Cost} &
\multirow{2}{*}{Time Taken (secs)} 
 \\ \cline{5-13} 
&& & Size &  w/o Design & w Design & \% Difference & w/o Design & w Design & \% Difference & w/o Design &  w Design & \% Difference &\\
\hline
\hline
\multirow{4}{*}{Blocksworld} &\multirow{2}{*}{1}& Avg & 1.25&14.11&	2.18&	-84.54 &	8.69&	9.52 &	9.58&	16.28&	4.87&-70.07& \multirow{4}{*}{1800}\\
\cline{3-13}
&& SD & 0.79 &	16.86 &0.92	&-&	1.39&	1.85	&-&	17.11&1.38&-&\\
\cline{2-13}
&\multirow{2}{*}{10}& Avg & 1.25&91.90 &	14.20 &	-84.54 &	8.69& 9.52 & 9.57&	113.63&38.33& -66.27 &\\
\cline{3-13}
&& SD & 0.78&	109.80&	5.98	&-&	1.39 &	1.85	&-&	112.36 &	9.59& -&\\
\hline
\hline
\multirow{4}{*}{IPC-Grid} &\multirow{2}{*}{1}& Avg & 0.75&3571.84 &	1455.39& 	-59.25&24.84&	24.84&	0 &23326.29 &	1461.79 &	-93.73& \multirow{4}{*}{1800}\\
\cline{3-13}
&& SD & 0.44&12043.62&4428.98&-&	3.01&	3.01	&-&	78444.61&4429.19&-&\\
\cline{2-13}
&\multirow{2}{*}{10}& Avg & 0.75&23264.19&	9479.32&	-59.25&	24.84&24.84&0&23326.29&	9541.61&	-59.09& \\
\cline{3-13}
&& SD & 0.44 &78442.72 &28846.93&-&	3.01&	3.01&-&	78444.61&28848.86& -&\\
\hline
\hline
\multirow{4}{*}{Driverlog} &\multirow{2}{*}{1}& Avg & 0.8 & 2.26	&1.6	&  - 29.14 & 8.46& 9.17  &	8.46 &	4.37&	4.09  &	- 6.39 & \multirow{4}{*}{219.42}\\
\cline{3-13}
&& SD & 0.77 & 0.54& 0.57  &-&	0.59&		0.89  &-& 0.61& 0.54 	&-&\\
\cline{2-13}
&\multirow{2}{*}{10}& Avg & 1.2&	14.70&	8.93  &	-39.28 &	8.45&	9.71  &	14.76 & 35.85& 33.50  &	- 6.57  & \\
\cline{3-13}
&& SD & 0.69&3.54& 2.78 & -&	0.59&	0.97  &-  &	4.30&	3.94& -&\\
\hline
\end{tabular}%
}
\caption{We report the impact of design modifications on inexplicability score, plan cost and total cost. We also report the average and standard deviation values for the three optimization terms in the objective function along with the run time.}
\label{table:eval1}
\end{table*}

\subsection{Performance on IPC domains} For this objective, we set
$\alpha$, $\beta$ and $\kappa$ to 1.0, 0.25, 0.25 respectively for all domains i.e., we gave more weight to minimizing inexplicability. We set $\mathcal{T}$ to 1 and 10 and $\gamma$ to 0.9. We allowed the search to run for 30 minutes per problem. If it ended within 30 minutes we output the optimal design modification, else we output the design modification which gave the best optimization value (or total cost) among the explored nodes. 
Note that in the \texttt{IPC grid}, we restricted the set of applicable designs at each node to the ones that affect the current expected human plan.
To show the impact of design modifications, we computted the inexplicability score, the plan cost, and the total cost for the most explicable plan in the initial model without any design modification. 
To compare the impact of longitudinality, we compute these for single step horizon and multi-step horizon. 

\begin{figure}[!t]
\centering
\includegraphics[width=0.8\columnwidth]{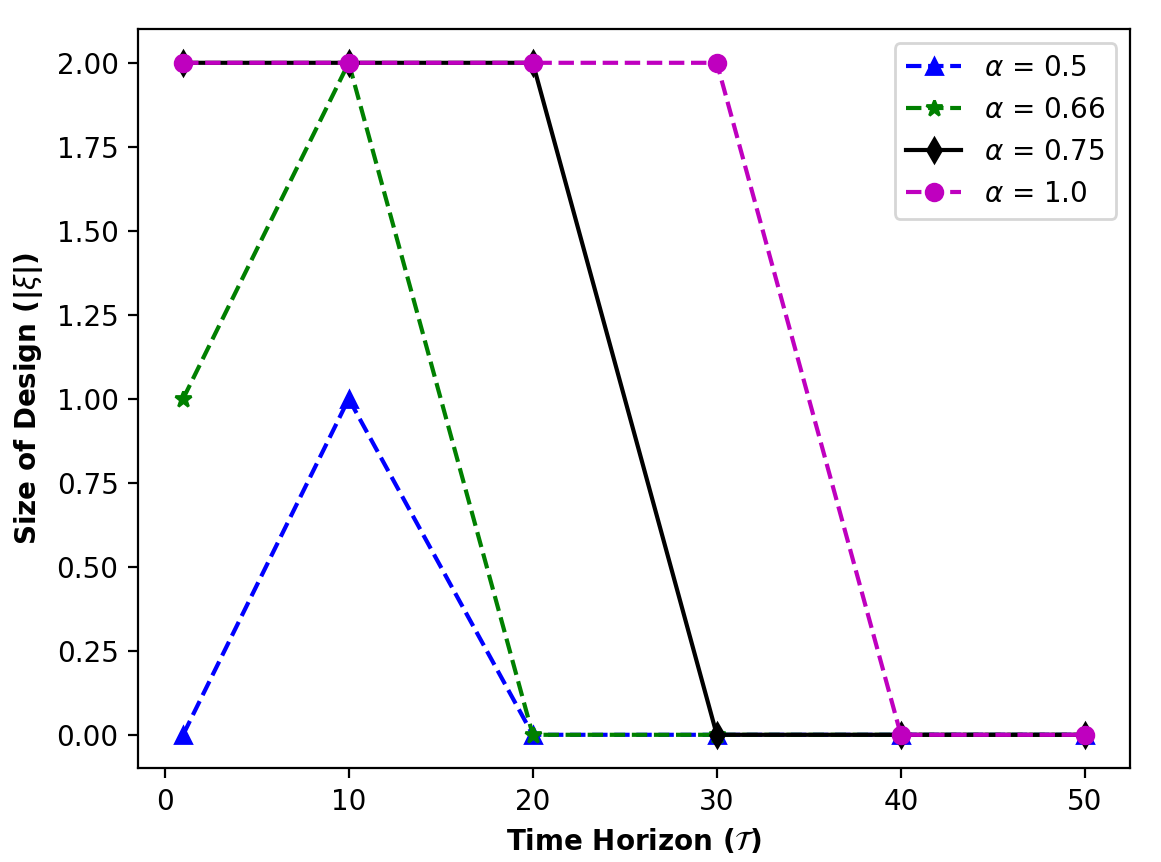}
\caption{The plot shows the impact of inexplicability score coefficient ($\alpha$) on design size in the solutions over different time horizons for a \texttt{Driverlog} problem.}
\label{fig:graph1}
\end{figure} 

In Table \ref{table:eval1}, we report the results for the 3 domains.
By comparing the inexplicability score with and without design, we see that the inexplicability always decreases as expected. For Blocksworld and IPC-Grid, the percentage decrease is the same for one-step and multi-step horizon; this is because the same set of designs were the best solutions found for both settings (under the time-limit) and the values got multiplied with the value of $\mathcal{T}$. On the other hand, for \texttt{Driverlog}, there were different designs selected, as is evident from the values. By comparing the plan cost with and without design, we can see that for \texttt{Blocksworld} and \texttt{Driverlog}, there is a substantial increase in the plan cost. This is because for these two domains, the designs ensured an action could be performed only after execution of another action. In this case, the robot bears additional cost for improving the explicability. On the other hand for \texttt{IPC-Grid}, the action pruning strategy removed actions from the human's mental model and therefore there is no increase in the plan cost. Similarly, by comparing the total cost with and without design, we can see that there is a significant decrease in the total cost after applying design modifications. This is because the optimization chooses design modifications that minimize the overall cost associated with the initial model.


\subsection{Interplay Between Inexplicability Score and Plan Cost}

To study the interplay between inexplicability score and plan cost, we experimented with a $\mathcal{DP}_{Exp,\mathcal{D}, \mathcal{T}}$ problem in the \texttt{Driverlog} domain. We used discount factor $\gamma = 0.9$ and design cost coefficient $\beta = 0.25$. We tested the impact of different inexplicability score coefficient values ($\alpha$: 0.5, 0.66, 0.75, 1) on the number of design choices in an optimal solution given different time horizons $\mathcal{T}$: 1, 10, 20, 30, 40, 50. At most two design choices were allowed in the solution. 

In Figure \ref{fig:graph1}, we report the impact on the size of design modifications. Recall that, the discount factor $\gamma$ denotes the probability that the human will not ignore the inexplicability of the behavior. Therefore, when $\gamma$ is set to 0.9, the optimization prioritizes reduction in inexplicability score. Given that the design cost coefficient $\beta = 0.25$ is low, even with single time step horizon $\mathcal{T} = 1$, designs are found that improve the explicability of the robot's behavior as shown in Figure \ref{fig:graph1}. However, the designs in the \texttt{Driverlog} domain lead to an increase in the cost of the robot plan (due to additional disembark actions). Given a long time horizon ($\mathcal{T} = 50$), the cost overhead borne by the robot for being explicable becomes greater than the impact of the inexplicability score on the human. Hence no designs are found at $\mathcal{T} = 50$ for any of the $\alpha$ values. If explicability of the robot's behavior is desired for longer horizons, this can be achieved by setting $\alpha$ to a high value. This shows the inherent interplay between the inexplicability of the behavior and the additional plan cost borne by the robot to reduce inexplicability.

\section{Related Work}

This work explores the connection between two parallel threads of active research: one on environment design and the other on explicable behavior. The problem of environment design is connected to that of mechanism design \cite{narahari2014game}, which has been thoroughly investigated by the game theory community. Environment design \cite{zhang2009general} involves modifying the environment so as to maximize or minimize some objective for an agent \cite{ijcai2017-608}. The problem of design has been leveraged to simplify related problems like goal recognition \cite{keren2014goal}, plan recognition \cite{mirsky2019goal}, etc. These works have studied the possibility of modifying the environment so as to make the robot's behavior easily recognizable. 
These works have also looked at various types of designs, including, action pruning \cite{keren2014goal}, action conditioning \cite{keren2018strong}, sensor refinement \cite{keren2018strong}, sensor placement \cite{keren2016privacy}, etc.
The problem of environment design has also been studied for stochastic actions \cite{wayllace2016goal,wayllace2017new}.

The notion of explicability was introduced in \cite{exp-yu}, which discussed generating explicable behavior by learning the sequence of actions that are explicable to the humans. \cite{explicable-anagha} explored the notion of explicability given knowledge of the human's mental model, and used plan distances as a stand in for $\delta_{\mathcal{P}_H}$. Generation of explicable behavior has also been studied in combination with explanations \cite{balancing}. Further, \cite{exact} explores the use of explanatory actions to convert the explanation generation problem to a sequential decision making problem. Moreover, \cite{landscape} explores the connections between explicability and other types of interpretable behaviors like legibility \cite{dragan2013legibility, unified-anagha}, and predictability \cite{dragan2013legibility,fisac2018generating}.

\section{Discussion and Conclusion}

In this paper, we bridge the gap between past works on environment design and those on generation of explicable behavior. We present a novel framework of environment design for explicability. The notion of environment design makes sense when there is repeated execution of tasks or when there are multiple tasks in the environment. 
This allows us to explore a novel trade-off that arises between one-time cost of design and the repeated cost overhead incurred by the robot for generating explicable behavior. 
In general, the design modifications can also be software changes that only affect the robot's capabilities.
In prior works on explicable plan generation, the underlying setting considered a one-time interaction between a human and a robot. In this work, we relaxed this assumption and explored the notion of inexplicability given repeated interactions. 

In this work, we assumed that the robot is capable of performing explicable behavior. However, we can also consider the problem of environment design for explicability when the robot is rational but not cooperative (i.e. it can only generate cost-optimal plans in the given environment and will not bear the overhead cost of being explicable). In this case, the emphasis is on choosing a set of design modifications which reduce the worst case inexplicability score associated with cost-optimal plans for a task. 
Similarly, we can also consider the problem of environment design for explicability when the robot can communicate (i.e. it generates cost-optimal plans but it provides an explanation to make its behavior explicable to the human). In settings involving different humans, the robot will have to provide the same explanation over and over to make its behavior explicable. Therefore, we again see similar trade-offs between one-time design cost versus the cost of repeated explanations borne by the robot. This would require modeling the impact of longitudinal interactions on explanations to account for how the human will update their mental model each time they receive an explanation. 

\section*{Acknowledgements}

The research of ASU contributors is supported in part by ONR grants N00014-16-1-2892, N00014-18-1-2442, N00014-18-1-2840, N00014-9-1-2119, AFOSR grant FA9550-18-1-0067, DARPA SAIL-ON grant W911NF-19-2-0006, NASA grant NNX17AD06G, and a JP Morgan AI Faculty Research grant.

\bibliographystyle{IEEEtran}
\bibliography{bib}

\end{document}